%% file: manuscript.tex
\documentclass[9pt,twocolumn,twoside]{pnas-new}
\templatetype{pnasresearcharticle} 
\include{packages}
\include{acronyms}
\newcommand{\ra}{7a}

\title{\mytitle}

\author[a,1]{Niklas T\"otsch}
\author[a]{Daniel Hoffmann} 

\affil[a]{Bioinformatics and Computational Biophysics, Universit\"at Duisburg-Essen, 45141 Essen, Germany}

\leadauthor{T\"otsch} 

\authorcontributions{N.T. and D.H. designed research; N.T. performed research and analyzed data; N.T. and D.H. wrote the paper.}
\authordeclaration{The authors declare no conflict of interest.}
\correspondingauthor{\textsuperscript{1}To whom correspondence should be addressed. E-mail: niklas.toetsch@uni-due.de}


\begin{abstract}
  Classifiers are often tested on relatively small data sets, which should lead to uncertain performance metrics.
  Nevertheless, these metrics are usually taken at face value.
  We present an approach to quantify the uncertainty of classification performance metrics, based on a probability model of the confusion matrix.
  Application of our approach to classifiers from the scientific literature and a classification competition shows that uncertainties can be surprisingly large and limit performance evaluation. 
  In fact, some published classifiers are likely to be misleading.
  The application of our approach is simple and requires only the \acrlong{cm}. It is agnostic of the underlying classifier. Our method can also be used for the estimation of sample sizes that achieve a desired precision of a performance metric.
\end{abstract}

\begin{document}

\maketitle
\thispagestyle{firststyle}
\ifthenelse{\boolean{shortarticle}}{\ifthenelse{\boolean{singlecolumn}}{\abscontentformatted}{\abscontent}}{}

Classifiers are ubiquitous in science and every aspect of life.
They can be based on experiments, simulations, mathematical models or even expert judgement.
The recent rise of machine learning has further increased their importance.
But machine learning practitioners are by far not the only ones who should be concerned by the quality of classifiers.
Classifiers are often used to make decisions with far-reaching consequences.
In medicine, a therapy might be chosen based on a prediction of treatment outcome.
In court, a defendant might be considered guilty or not based on forensic tests.
Therefore, it is crucial to assess how well classifiers work.

In a binary classification task, results are presented in a 2$\times$2 \gls{cm}, comprising the numbers of \gls{tp}, \gls{fn}, \gls{tn} and \gls{fp} predictions.

\begin{equation}
\acrshort{cm} = \begin{bmatrix} \acrshort{tp} & \acrshort{fn} \\ \acrshort{fp} & \acrshort{tn} \end{bmatrix}
\label{eq:cm}
\end{equation}

\Gls{cm} contains all necessary information to determine metrics which are used to evaluate the performance of a classifier.
Popular examples are \gls{acc}, \gls{tpr}, and \gls{tnr} 

\begin{align}
\acrshort{acc} &= \frac{\acrshort{tp} + \acrshort{tn}}{\acrshort{tp} + \acrshort{fn} + \acrshort{fp} + \acrshort{tn}} \\
\acrshort{tpr} &= \frac{\acrshort{tp}}{\acrshort{tp} + \acrshort{fn}} \\
\acrshort{tnr} &= \frac{\acrshort{tn}}{\acrshort{tn} + \acrshort{fp}}
\end{align}
These are given as precise numbers, irrespective of the \glspl{n} used for their calculation in performance tests. 
This is problematic especially in fields such as biology or medicine, where data collection is often expensive, tedious, or limited by ethical concerns, leading often to small \glspl{n}. 
In this study we demonstrate that in those cases the uncertainty of the \gls{cm} entries cannot be neglected, which in turn makes all performance metrics derived from the \gls{cm} uncertain, too. 
In the light of the ongoing replication crisis~\cite{baker2016reproducibility}, it is plausible that negligence of the \acrlong{mu} impedes reproducible classification experiments.

There is a lack of awareness of this problem, especially outside the machine learning community.
One often encounters discussions of classifier performance lacking any statistical analysis of the validity in the literature. 
If there is a statistical analysis it usually relies on frequentist methods such as confidence intervals for the metrics or \gls{nhst} to determine if a classifier is truly better than random guessing.
\Gls{nhst} ``must be viewed as approximate, heuristic tests, rather than as rigorously correct statistical methods''~\cite{Dietterich1997}.

Bayesian methods can be valuable alternatives.~\cite{Benavoli2017} 
To properly account for the uncertainty, we have to replace the point estimates in the \gls{cm} and all dependent performance metrics by probability distributions.
Correct and incorrect classifications are outcomes of a Binomial experiment.~\cite{Brodersen2010}
Therefore, Brodersen et al. model \gls{acc} with a \gls{bbd}

\begin{equation}
\acrshort{acc} \sim \mathrm{Beta}(\acrshort{tp} + \acrshort{tn} + 1, \acrshort{fp} + \acrshort{fn} + 1).
\label{eq:acc_bbd}
\end{equation}
Some of the more complex metrics, such as balanced accuracy, can be described by combining two \glspl{bbd}.~\cite{Brodersen2010}

Caelen presented a Bayesian interpretation of the \gls{cm}.~\cite{Caelen2017}
This elegant approach, based on a single Dirichlet-multinomial distribution, allows to replace the count data of the confusion matrix with distributions which account for the uncertainty.

\begin{align}
\acrshort{cm} &\sim \mathrm{Mult}(\acrshort{cpm}, \acrshort{n}) \label{eq:mult} \\
\acrshort{cpm} &\sim \mathrm{Dirichlet}((1, 1, 1, 1))
\label{eq:dcm}
\end{align}
where \acrshort{cpm}=$\left[ \parm{\acrshort{tp}}, \parm{\acrshort{fn}}, \parm{\acrshort{tn}}, \parm{\acrshort{fp}} \right]$ is the \acrlong{cpm} which represents the probabilities to draw each entry of the \gls{cm}.
The major advantage of Caelen's approach over the one presented by Brodersen lies in a complete description of the \gls{cm}.
From there, all metrics can be computed directly, even those that cannot simply be described as \gls{bbd}.

Caelen calculates metric distributions from confusion matrices that are sampled according to Equation~\ref{eq:mult}.
Here, we demonstrate that this approach is flawed and derive a correct model.
Whereas previous studies focused on the statistical methods, we prove that classifier performance in many peer-reviewed publications is highly uncertain.
We studied a variety of classifiers from the chemical, biological and medicinal literature and found cases where it is not clear if the classifier is better than random guessing.
Additionally, we investigate \acrlong{mu} in a Kaggle machine learning competition where \acrlong{n} is relatively large but a precise estimate of the metrics is required.
In order to help non-statisticians to deal with these problems in the future, we derive a rule for sample size determination and offer a free, simple to use webtool to determine \acrlong{mu}.

\section{Methods}

\subsection{Model}

The \gls{cpm}, that is the probabilities to generate entries of a confusion matrix, can be derived if \gls{prev}, \gls{tpr} and \gls{tnr} are known.~\cite{Kruschke2015c}

\begin{align}
\parm{\acrshort{tp}} &= \acrshort{tpr} \cdot \acrshort{prev} \\
\parm{\acrshort{fn}} &= (1 - \acrshort{tpr}) \cdot \acrshort{prev} \\
\parm{\acrshort{tn}} &= \acrshort{tnr} \cdot ( 1- \acrshort{prev}) \\
\parm{\acrshort{fp}} &= (1 - \acrshort{tnr}) \cdot ( 1- \acrshort{prev})
\end{align}

\begin{figure}
\centering
\includegraphics[width=\columnwidth]{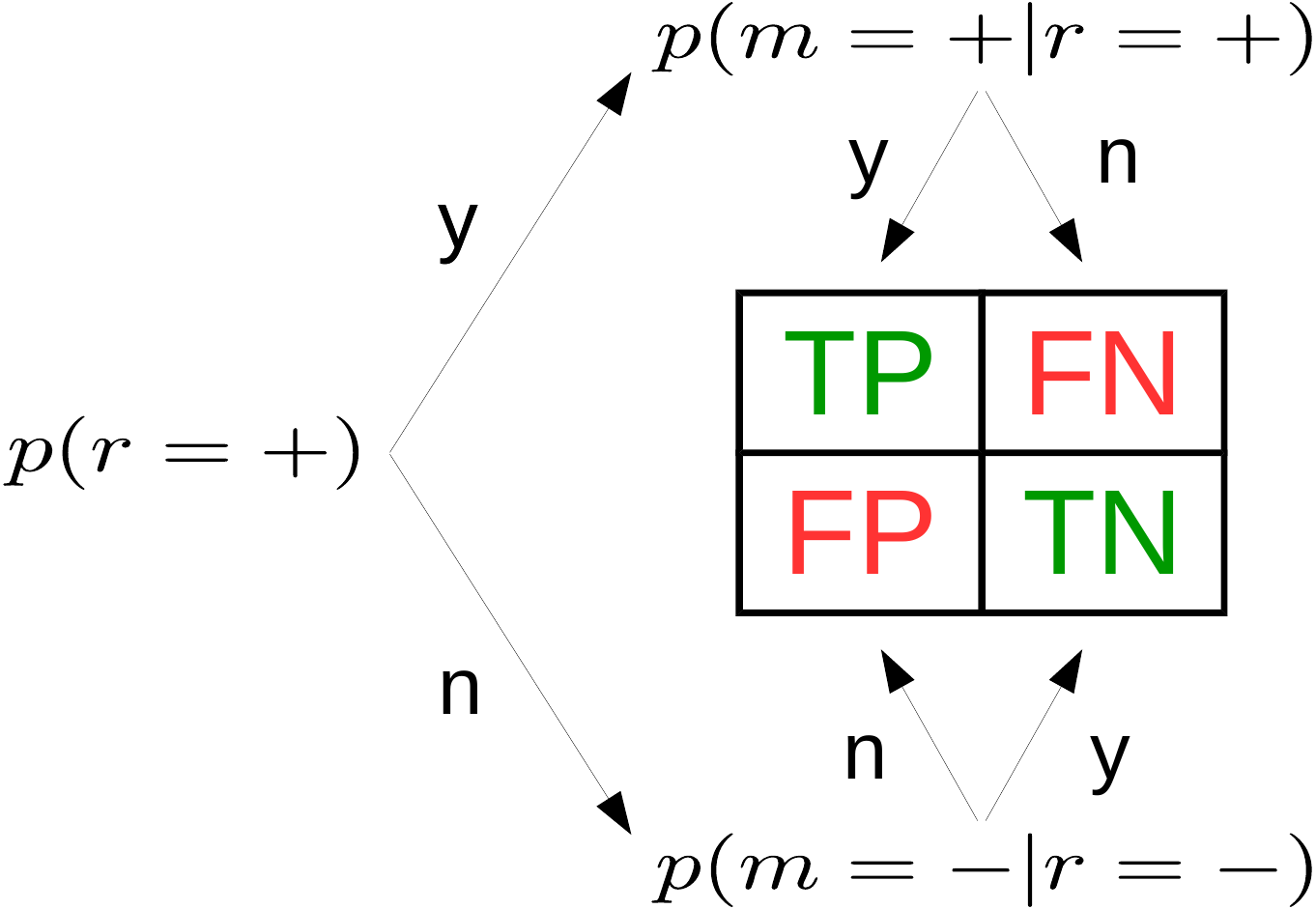}
\caption{Three \acrlongpl{bbd} $p(\cdot)$ -- \acrlong{prev} (left), \acrlong{tpr} (top), \acrlong{tnr} (bottom) -- define the \acrlong{cm}.
Based on them, all entries of the \gls{cm} can be expressed as distributions with explicit uncertainty due to limited sample size.}
\label{fig:scheme}
\end{figure}

The idea that these metrics can also be inferred from data, propagating the uncertainty, is the starting point of the present study.
Using three \glspl{bbd}, one for each of \gls{prev}, \gls{tpr} and \gls{tnr}, we can express all entries of the \gls{cm} (Figure~\ref{fig:scheme}).
Since \gls{prev}, \gls{tpr} and \gls{tnr} are distributions, the entries of  \gls{cpm} $\left[ \parm{\acrshort{tp}}, \parm{\acrshort{fn}}, \parm{\acrshort{tn}}, \parm{\acrshort{fp}} \right]$ are too.
Based on \gls{cpm} we calculate all other metrics of interest.

For the following Bayesian treatment we use the Laplace prior, $\mathrm{Beta}(\alpha=1, \beta=1)$, for \gls{prev}, \gls{tpr} and \gls{tnr} because its uniform distribution introduces no bias, which makes it suitable for any classification problem.
It is noteworthy that a flat prior on \gls{prev}, \gls{tpr} and \gls{tnr} leads to non-flat priors on other metrics (Section~S1).
We discuss two additional objective priors in the supplementary material.
If additional knowledge is available, based e.g.\ on the experimental setup of the classifier, it should be incorporated in the prior. 
Here, we refrain from using informative priors to keep the method generally applicable.
 
Our approach is quite similar to Caelen's but has distinct advantages.
First, \gls{prev}, \gls{tpr} and \gls{tnr} are common metrics; thus prior selection is easier.
Second, our model clearly distinguishes data intrinsic \gls{prev} from the classifier intrinsic measures \gls{tpr} and \gls{tnr}.
Consequently, our approach allows to ``exchange'' \gls{prev}.
This is useful if the prevalence of the test set differs from the prevalence of the population the classifier will be applied to in production.
Such a scenario is common in medical tests where \gls{prev} is very low in the general population.
To increase the sample size of positive cases in the test set without inflating the number of negative ones, \gls{prev} differs from the general population.
Using a Dirichlet-multinomial distribution, it is not straightforward to evaluate a classifier for a different \gls{prev}.
If the data set was designed to contain a specified fraction of positive and negative instances, \gls{prev} is known exactly (Section~S2).
This scenario is easy to implement in our model but not in Caelen's.

Depending on the context, \gls{prev} may have two meanings.
If one is interested in a population, \gls{prev} describes how common fulfilment of the positive criterion is.
For an individual, e.g.\ a patient, \gls{prev} can be considered the prior.
If additional information was available for this subject, such as results of previous tests,  \gls{prev} would differ from the \acrlong{prev} in the general population.
This prior can be updated with \gls{tpr} and \gls{tnr}, representing the likelihood, to yield the posterior for the individual.

\subsection{Measuring true rather than empirical uncertainty}

Bayesian models allow posterior predictions.
In our case, posterior predictions would be synthetic confusion matrices \cm{}, which can be generated from a multinomial distribution (Equation~\ref{eq:mult}).

\begin{figure}[!htb]
\centering
\includegraphics[width=\columnwidth]{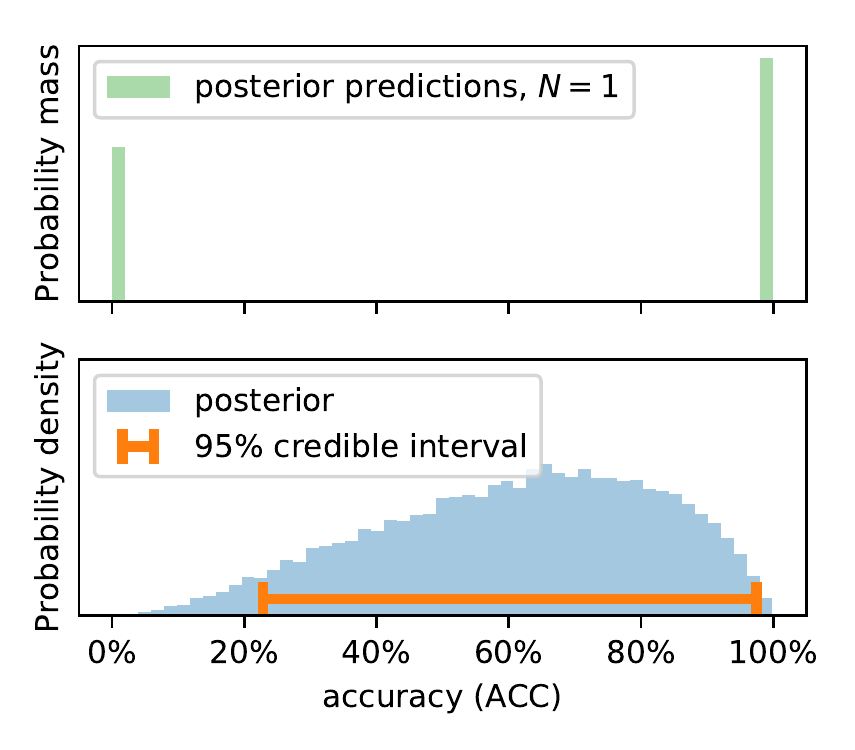}
\caption{Calculating \acrfull{acc} on posterior predictions of the  \acrlong{cm} yields a discrete distribution (top), representing expected observations of the metric at given \acrfull{n}.
Posterior distributions  (bottom) of the metric must be calculated from the inferred entries of the \acrfull{cpm} as outlined in the text.}
\label{figure:theta_vs_pp_metric}
\end{figure}

This approach is equivalent to a combination of two/three binomial distributions as shown in Figure~\ref{fig:scheme} but slightly more elegant for posterior predictions.
Caelen samples many \cm{} to obtain metric distributions, which requires a choice of sample size \gls{n}. 
Caelen uses the \gls{n} of the original \gls{cm} the parameters have been inferred from.
This is not satisfying because in this way only the empirical distribution of the metrics for a given \gls{n} is generated, not the true distribution of the metrics.
Consider the example of \gls{cm} = (\acrshort{tp}, \acrshort{tn}, \acrshort{fp}, \acrshort{fn}) = (1, 0, 0, 0), i.e.\ \mbox{\gls{n} = 1}.
We will consider this classifier's \gls{acc}.
Caelen's approach leads to a discrete distribution of the accuracy allowing only 0 and 1 (Figure~\ref{figure:theta_vs_pp_metric}, top).
There was one correct prediction in the original \gls{cm}, therefore it is impossible that the accuracy is 0. 
In other words, the probability mass at \gls{acc}=0 should be strictly 0.
If one is interested in the true continuous posterior distribution of a metric, one must calculate it from \gls{cpm} directly (Figure~\ref{figure:theta_vs_pp_metric}, bottom).
We prove in Section~S4 that Caelen's approach systematically overestimates the variance in metric distributions.

We still consider Caelen's way of calculating metrics extremely useful since it allows to tackle the problem of reproducibility.
Generating synthetic \cm{} according to Equation~\ref{eq:mult} allows us to estimate what would happen if multiple researchers applied the same classifier to different data sets of size \gls{n} and reported the corresponding \glspl{cm} and metrics.
Figure~\ref{figure:theta_vs_pp_metric} shows that 
they might report completely different values of a metric if \gls{n} is small.
Under these circumstances, classification experiments are not reproducible.

\subsection{\Acrlong{mu} equals credible interval length}

If there is little data available, posterior distributions are broad.
We define \gls{mu} as the length of the 95\% highest posterior density interval (``credible interval'').
There is a 95\% likelihood that the metric is within this credible interval (bottom of Figure~\ref{figure:theta_vs_pp_metric}).

\subsection{Implementation}

Since the beta distribution is the conjugate prior of the binomial distribution, the posterior distribution can be derived analytically.
There is no need for Markov chain Monte Carlo sampling.
This is merely a convenience, our approach would work with any prior.
To calculate metrics, we sampled 20000 data points.
Splitting these data points into two arrays of equal length, we use \texttt{PyMC}'s implementation of the Gelman-Rubin diagnostics ($R_c < 1.01$) to verify that the posterior distribution is properly sampled.~\cite{Gelman1992,Brooks1998,Salvatier2016}

The implementation of our model in \texttt{Python} can be found on \url{https://github.com/niklastoe/classifier_metric_uncertainty}.

\section{Results and Discussion}

\subsection{Classifier examples from the literature}

To assess the uncertainty in classifier performance in the scientific literature, we searched \texttt{Google Images} for binary confusion matrices from peer reviewed publications in the area of chemistry, biology and medicine with less than 500 samples in the test set.
We collected 24 classifiers; confusion matrices and the references to the publications are listed in Table~S1.
Publications are indexed with numbers. 
If more than one classifier is presented in one publication, a character is added.  
Some of these classifiers are based on statistical models of available data. 
Others are based on simulations.
The majority of publications describe the development of a new experimental approach followed by a statistical model that transforms the experimental outcome into a classification.
Classifiers come from diverse fields, e.g. chemical detection (adulterants in palm oil or cocaine, mycotoxins in cereals) or prediction of inhibitors of amyloid-aggregation or enzymes.
The smallest sample size was 8, the largest 350.

While the resources invested in the development of these classifiers must have been considerable, their performance had not been thoroughly evaluated. Specifically, only for a single classifier the uncertainty had been quantified by calculating confidence intervals.
In some of the literature examples, we also noted severe problems unrelated to small \gls{n}.
Due to usage of \gls{acc} for imbalanced data sets and mixing of train and test data sets for reported metrics, the performance of some classifiers was overrated.
These problems have been addressed previously.~\cite{chicco2017} 
In this study, we evaluate classifiers on metrics which are invariant to class imbalance and rely exclusively on test data sets.

 Our selection may not in all aspects be representative of published classifiers in any field. However, the negligence of metric uncertainty observed in this selection is not exceptional.
Our choice of biology, chemistry, and medicine as scientific domain was based on our relative familiarity with those fields. While in this domain small sample sizes are common (due to costly data collection), this problem is probably not limited to this domain.

\subsection{Metrics are broadly distributed}

\begin{figure*}
\centering
\includegraphics[width=6.7in]{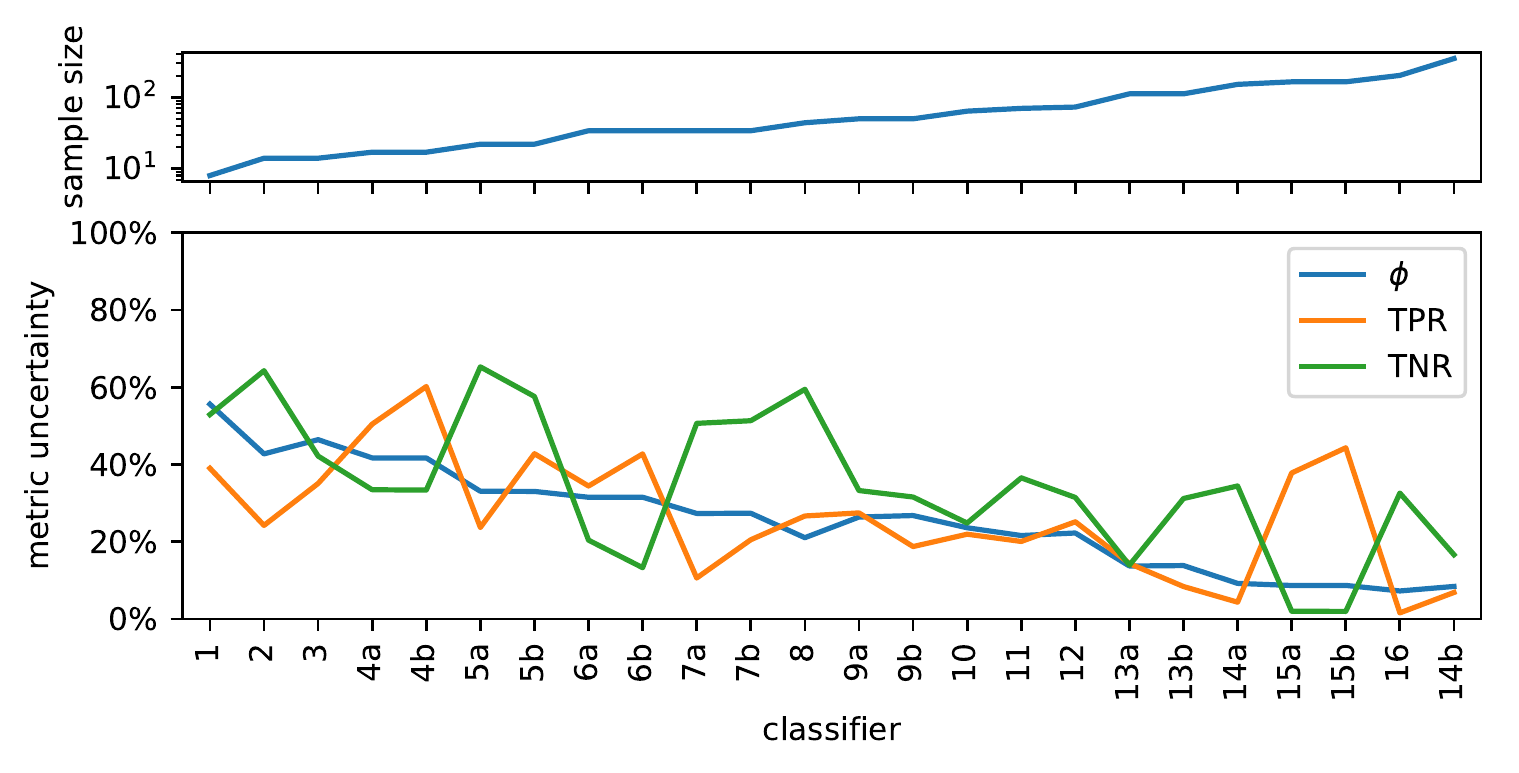}
\caption{\Acrfull{mu} for \acrfull{prev}, \acrfull{tpr}, and \acrfull{tnr} of literature examples. 
Classifiers are sorted by sample size (top).
\acrshort{mu} is large but decreases with sample size.
Since \acrshort{mu} is determined by the length of the 95\% highest posterior density interval, the theoretical upper limit is 95\% (in which case little is known about the metric).
If \acrshort{mu} was 0\%, the corresponding metric would be known at infinite precision.}
\label{figure:ci_lengths}
\end{figure*}

Typically, classifier metrics are reported as single numerical values (often to one or more decimals) without indication of uncertainty.
However, the true \glspl{mu} of classifiers in our collection are too large to be ignored (Figure~\ref{figure:ci_lengths}).
Often, \gls{mu} is greater than 20~percentage~points, sometimes exceeding 60~percentage~points.
In general, \gls{mu} in all three observed metrics declines as \gls{n} increases. 
The decrease is not monotonous because \gls{mu} also depends on the value of the metric (Section~S5).

The \glspl{mu} we show in Figure~\ref{figure:ci_lengths} were obtained from \gls{cpm}.
As mentioned above, metrics calculated from empirically observed confusion matrices of the same classifier would vary even more. Thus, if an independent lab tried to reproduce \gls{cm} for, say, example \ra{}, with a much larger sample size, \gls{tnr} values of 90\% or 50\% would not be surprising, although the value given in the paper is 75\%.

It is possible that we underrate some classifiers.
If a metric should have a more informative prior than the Laplace prior we used, e.g.\ due to previous experience or convincing theoretical foundations, the posterior could also be more narrowly defined.

\subsubsection{Metric uncertainty limits confidence in high-stakes application of classifiers}

In the following, we discuss in greater detail  \gls{mu} for one classifier where the consequences of misclassification are dramatic and understandable to non-experts.
Classifier \ra{} is a new method to predict cocaine purity based on a "simple, rapid and non-destructive" experiment followed by mathematical analysis.
The authors stress the importance of such a method for forensic experts and criminal investigators.
Predictions are compared to a destructive and more elaborate experimental reference.
Prosecutors in countries such as Spain may consider purity as evidence of the intent to traffic a drug, presumably resulting in more severe punishments.\footnote{http://www.emcdda.europa.eu/system/files/publications/3573/Trafficking-penalties.pdf; accessed December 3rd, 2019}
Consequently, a \gls{fp} would result in a wrongful charge or conviction causing severe stress and eventually imprisonment for the accused.
A \gls{fn} on the other hand might lead to an inadequately mild sentence.
Moreover, one could also consider the scenario of drug checking.
In some cities, such as Zurich, Switzerland, social services offer to analyze drugs to prevent harm from substance abuse due to unexpectedly high purity or toxic cutting agents.\footnote{\url{https://www.saferparty.ch/worum-gehts.html}; accessed on June 9th, 2020 at 3:42~pm}
In this context, a \gls{fn} could lead to an overdose due to the underestimated purity.

\begin{figure}
     \centering
     \begin{subfigure}[b]{0.7\columnwidth}
         \centering
         \include{rodrigues_cm}
         \caption{\Acrlong{cm}, r stands for reference, m for model}
         \label{fig:rodrigues2013_cm}
     \end{subfigure}
     \\
     \begin{subfigure}[b]{\columnwidth}
         \centering
         \includegraphics[width=\textwidth]{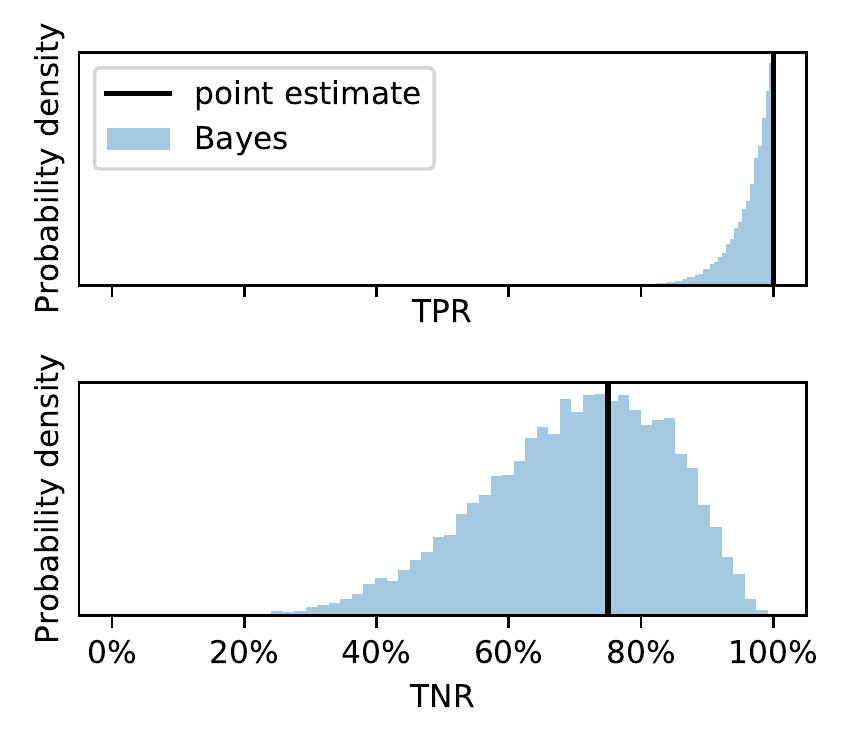}
         \caption{Posterior distributions for \acrfull{tpr} (top) and \acrfull{tnr} (bottom)}
         \label{fig:rodrigues2013_posterior}
     \end{subfigure}
        \caption{Metric uncertainty for cocaine purity classifier \ra{}}
        \label{fig:rodrigues2013}
\end{figure}

The confusion matrix in Figure~\ref{fig:rodrigues2013_cm} is transcribed from the original publication.
We do not know whether their method was used for drug checking or in court (at least the authors received the samples from the local police department). If it was, could it be trusted by a forensic expert, judge, or member of the jury?
The posterior distribution of the \gls{tpr} (Figure~\ref{fig:rodrigues2013_posterior}) answers this question probabilistically.
The point estimate from \gls{cm} would be \gls{tpr}=100\% but due to small \gls{n}, the uncertainty is large.
The credible interval spans from 89\% to almost 100\% although not a single \gls{fn} has been observed in the test set.

Now consider \gls{tnr}.
Since there are only eight low purity cocaine samples, the uncertainty is much larger. 
While the point estimate would be \acrshort{tnr}=75\%, the credible interval is 43\%-95\%. 
It is possible, although unlikely, that the classifier would generate more \gls{fp} than \gls{tn}.
This would translate into more wrongful convictions than correct acquittals for possessing cocaine with high purity if this method was used as main evidence in court.

Our approach would hopefully lead to more cautious use of little tested classifiers.
Imagine two scenarios.
In the first, a judge is told that the forensic method has a \gls{tpr} of 100\% and a \gls{tnr} of 75\%.
In the second, she is told that it has an estimated \gls{tpr} of 89-100\% and an estimated \gls{tnr} of 43-95\%.
In the latter, the judge would be more hesitant to base her verdict on the classifier.

We do not know if \gls{prev} in the test set is representative of the prevalence of drug samples in criminal cases. Therefore, we cannot reasonably estimate the distribution of probabilities of wrongfully harsh/lax sentences. 
For a meaningful assessment of evidence, both \gls{prev} and \gls{mu} should be taken into account.
Our approach facilitates such an analysis.

\subsection{Some published classifiers might be deceptive}

As classification problems vary greatly so does the relevance of different metrics, depending on whether \gls{fn} or \gls{fp} are more or less acceptable.
Often, classifier development requires a tradeoff between \gls{fn} or \gls{fp}.
In this respect, \gls{bm} is of interest because it combines both in a single metric without weighting and measures the probability of an informed prediction.~\cite{Powers2011}

\begin{equation}
\acrshort{bm} = \acrshort{tpr} + \acrshort{tnr} - 100\%
\label{eq:bm}
\end{equation}

\begin{figure*}
\centering
\includegraphics[width=6.7in]{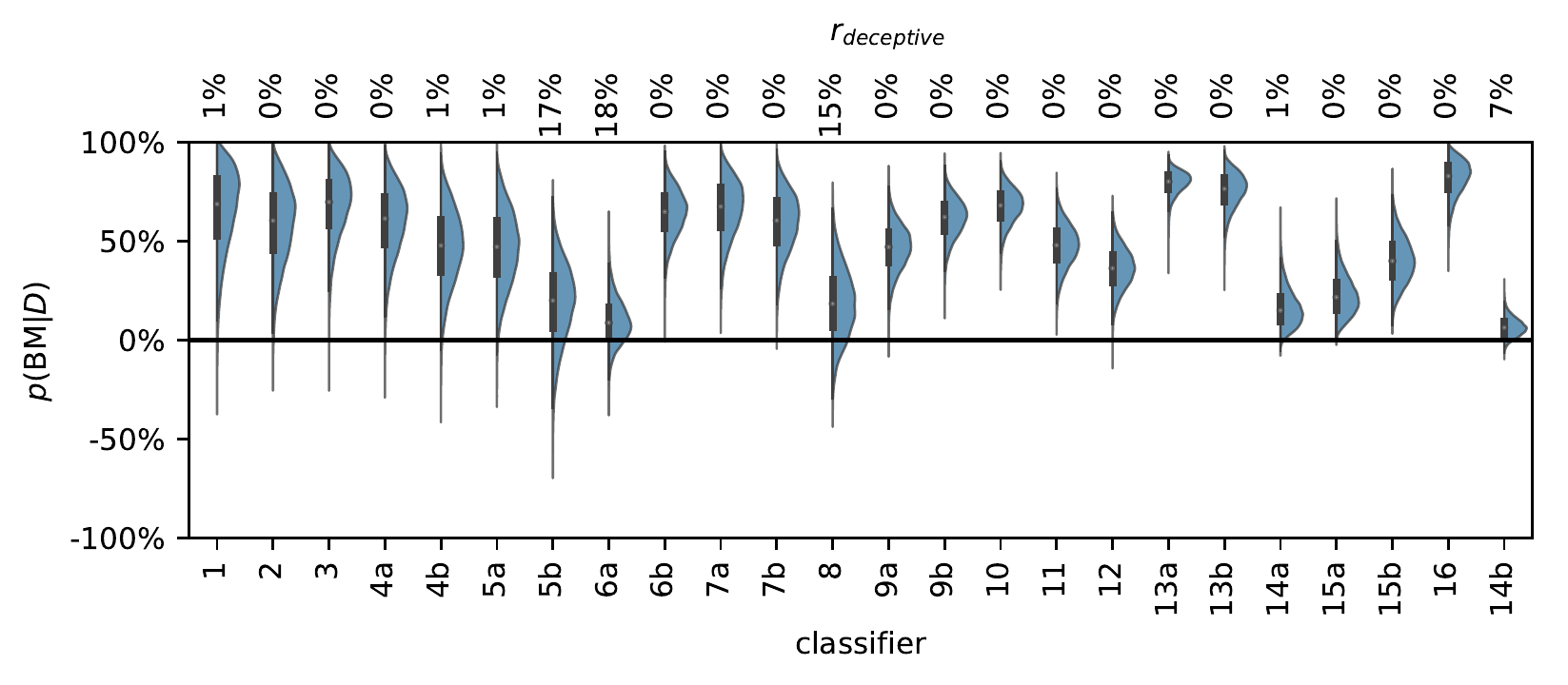}
\caption{Posterior distributions of \acrfull{bm} are broad due to small test sets in the literature examples.
Classifiers are sorted by ascending sample size as in Figure~\ref{figure:ci_lengths}.
Some classifiers have considerable posterior density in the negative region; these classifiers could be misinformative.
Percentages along top margin are \gls{rdec} values (Equation~\ref{eq:classifier_probabilities}), the probability that a classifier is worse than random guessing.}
\label{fig:literature_BM}
\end{figure*}

If \gls{bm}=100\%, prediction is perfect and the classifier is fully informed.
\gls{bm}=0\% means that the classifier is no better than random guessing and \gls{bm}=-100\% shows total disagreement, i.e.\ the predictor is wrong every single time.
Figure~\ref{fig:literature_BM} shows the posterior distributions of \gls{bm} for the collected examples from literature.
Due to small \gls{n}, they are broad.
Therefore, it is uncertain how much better the classifiers are compared to random guessing.
Several classifiers have considerable probability density in the negative region, i.e.\ it is possible that they are weakly deceptive.

We define the probabilities that a given classifier is informative or deceptive

\begin{align}
\acrshort{rinf} &= \int_{0\%}^{100\%} p(\acrshort{bm} | D) d \acrshort{cpm} \label{eq:rinf}\\
\acrshort{rdec} &= \int_{-100\%}^{0\%} p(\acrshort{bm} | D) d \acrshort{cpm}.
\label{eq:classifier_probabilities}
\end{align}

We determined \gls{rdec} for all literature examples (Figure~\ref{fig:literature_BM}, top).
Four classifiers have a considerable chance to be  deceptive.  
We note that three of them were published alongside alternative classifiers that the respective authors considered preferable (5b, 6a, 14b).
The probability that the classifier 8 is deceptive is approximately 15\% so we recommend to reevaluate it with a larger test set. 

The split of the BM posterior into \gls{rinf} and \gls{rdec} in Equations~\ref{eq:rinf} and \ref{eq:classifier_probabilities} is a coarse-graining device to ease conversation.
A classifier with a very low absolute \gls{bm} is neither informative nor deceptive but uninformative.

For finite \gls{n}, \gls{rdec} will be always greater than zero. What value of \gls{rdec} can be tolerated will of course depend on the application scenario, and should be carefully considered by developers and users of classifiers.

\subsection{Large \gls{n}, small difference in performance in metaanalysis of classifiers in machine learning}

Our approach can also be used for meta-analyses of classifier ensembles, an application that is of considerable interest in machine learning.~\cite{Dietterich1997, Benavoli2017, Calvo2019}
Kaggle, a popular online community for machine learning challenges, provides a suitable environment for such meta-analyses. 
On Kaggle, participants build classifiers and submit their results online to be evaluated and compared to those of others.
The best results are rewarded with cash prizes.
The metric for evaluation depends on the individual challenge.
Often, the competition is fierce and submitted results close, e.g.\ accuracy sometimes differs by less than one per mille.
With hundreds to tens of thousands of data points, test sets tend to be larger than in our literature collection above, but are still finite. 
Classifier metrics therefore retain some uncertainty, and statistical flukes could produce apparent differences in classifier performances that decide a competition.

\begin{figure}
\centering
\includegraphics[width=\columnwidth]{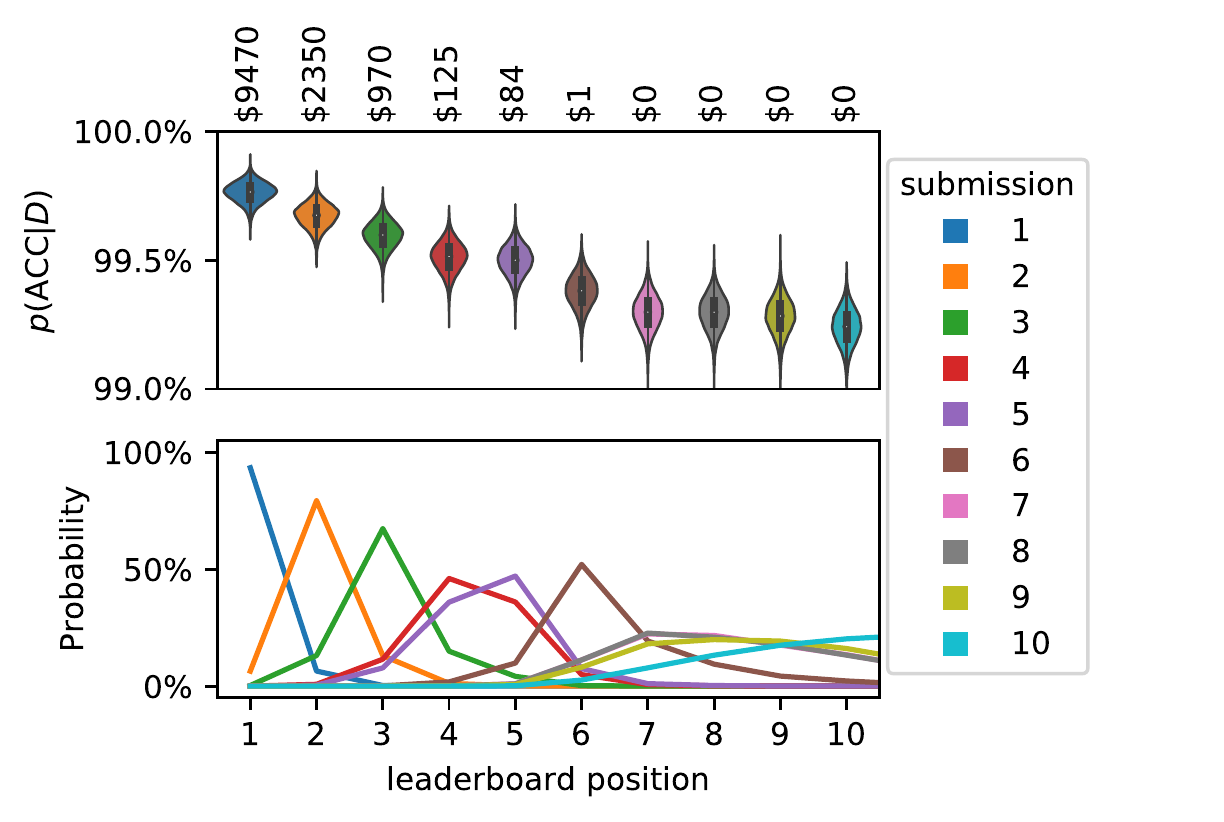}
\caption{\Acrfull{acc} posterior distribution for top ten submissions on Kaggle leaderboard (top). 
Distributions are narrow but the classifiers perform similarly. 
Therefore, after consideration of the uncertainty in \gls{acc}, the leaderboard positions of the submissions are uncertain (bottom).
If the cash prizes were awarded based on the probabilistic leaderboard, submissions outside of the top three would receive money (annotation).
These estimates, too, are uncertain by a few percentage points.}
\label{fig:probabilistic_leaderboard}
\end{figure}

We studied the Recursion Cellular Image Classification competition in greater detail.\footnote{\url{https://www.kaggle.com/c/recursion-cellular-image-classification/overview}; accessed on January 31st, 2020 at 9:25 am} 
Participants are tasked to properly classify biological signals in cellular images, disentangling them from experimental noise.
Submissions were ranked based on multiclass accuracy.
Micro-averaged multiclass accuracy can be modeled according to Equation~\ref{eq:acc_bbd}.
We evaluated private leaderboards, i.e.\ rankings provided by Kaggle with information on the participants and accuracies of their classifiers.
These private leaderboards were also used to award prizes.
Kaggle did not publish the exact size of the private test set but the overall test set contains 19899 images and the private leaderboards were calculated on approximately 76\% of it so we assumed \gls{n}=15123.
Based on \gls{n} and the published point estimates of \gls{acc} we could calculate \gls{tp}+\gls{tn} and \gls{fp}+\gls{fn} for every submitted classifier and compute a posterior distribution for \gls{acc} according to Equation~\ref{eq:acc_bbd} (Figure~\ref{fig:probabilistic_leaderboard}, top).

These posterior distributions overlap.
Using a Monte Carlo approach, we generated synthetic leaderboards from samples of the posterior distributions.
Counting how often every submission occurred at any leaderboard position yielded a probabilistic leaderboard (Figure~\ref{fig:probabilistic_leaderboard}, bottom).
We observed that the winning submission has a 93\% chance of being truly better than any other submission.
For leaderboard position 4 and worse, rank uncertainty becomes considerable and ranking validity is limited by the sample size.

At the end of this competition, the top 3 submission were awarded \$10.000, \$2.000 and \$1.000, respectively.
This implies that it is certain that the submissions listed in the top 3 positions are indeed the best classifiers.
As we have demonstrated, it is not certain which submissions are the best.
If one would weigh the awarded prizes based on the probability of a submission to be in each rank, other participants would have been awarded small cash prizes (Figure~\ref{fig:probabilistic_leaderboard}, top annotation).

Our approach is complementary to the Bayesian Plackett-Luce model, which considers multiple rankings for individual problems.~\cite{Calvo2019}
That model is agnostic about the performance metric since it is based only on the leaderboard position in every scenario. 
Consequently, it neglects the magnitude of the performance difference.
Our approach on the other hand requires a generative model for the performance metric but works for individual problems and quantifies the performance gap between classifiers.

\subsection{Sample size determination}

Since uncertainty in any commonly used metric decreases with increasing sample size \gls{n}, we can employ our approach of uncertainty quantification also to determine in advance values of \gls{n} so that a classifier fulfills predefined \gls{mu} criteria.

\begin{figure}
\centering
\includegraphics[width=\columnwidth]{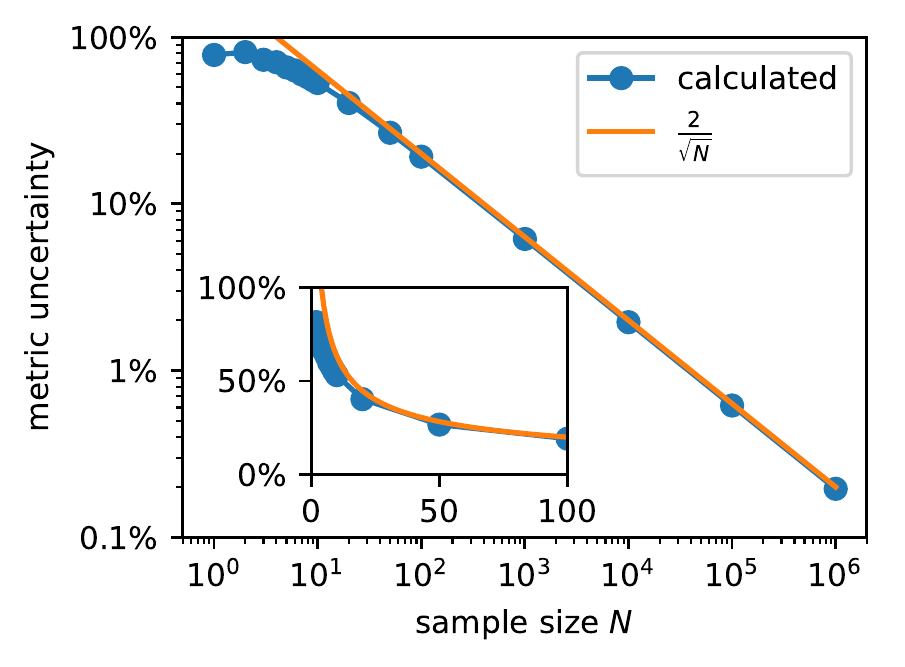}
\caption{Sample size determines \acrlong{mu} (defined by the length of the 95\% highest posterior density interval) for any metric whose distribution follows a \gls{bbd}.
Statistical power is 95\%.
The inset shows the same data on a non-logarithmic scale.}
\label{figure:sample_size_mu}
\end{figure}

For those metrics which can be described as \gls{bbd} (Equation~\ref{eq:acc_bbd}), such as \gls{acc}, \gls{tpr}, \gls{tnr} and \gls{prev}, we tested \gls{n} values spanning six orders of magnitude (Figure~\ref{figure:sample_size_mu}), following  Kruschke's protocol for sample size determination.~\cite{Kruschke2015l}
The shown results were obtained for a generating mode $\omega$=0.8 and concentration $k=10$.
We found that different $\omega$ yielded almost indistinguishable results at low $k$.

The probability to achieve a \gls{mu} more narrow than the given width in an empirical study, i.e.\ statistical power, is 95\%.
The interpretation is as follows:
If \gls{n}=100, the likelihood that $\gls{mu} \leq 19$ percentage points is 95\%.
In order to decrease \gls{mu} further, \gls{n} must be increased substantially.

Based on the standard deviation of a beta distribution and the central limit theorem we derive
\begin{equation}
\acrshort{mu} \lessapprox \frac{2}{\sqrt{\acrshort{n}}}
\label{eq:ssd_approximation}
\end{equation}
for $\acrshort{n} > 20$ in Section~S5.
It yields the correct order of magnitude which tells us if a classification study is feasible at the desired level of \gls{mu}.
This general rule ignores prior knowledge about the classifier.
The posterior of the metric derived from exploratory classification experiments should be considered.

We found several papers presenting metrics with one or even two decimals.
Classifier evaluations should be considered like any other experiment, and only significant digits should be given in their discussion.
Equation~\ref{eq:ssd_approximation} predicts that metric uncertainty would only drop below 0.1\%, which is necessary to present a metric with a decimal, if the test data set included several million data points.
Curating such a large test set is out of the question for the publications in our examples.
On Kaggle leaderboards, \gls{acc} is presented as percentage with three decimals.
Reducing metric uncertainty below 0.001\% would require tens of billions of data points.

\section{Conclusions}

In this work, we have presented a Bayesian model that quantifies the \acrlong{mu} of classifiers due to finite test sets.
It is completely agnostic about the underlying classifier.
Unlike previous work, our method cleanly separates data intrinsic \gls{prev} from classifier intrinsic \gls{tpr} and \gls{tnr}, which facilitates transfer to different data sets.
Nevertheless, our approach allows to evaluate  \acrlong{mu} of all metrics that are based on the \gls{cm}.

Our study of published examples suggests that \gls{mu} is a neglected problem in classifier development.
We found classifier metrics that were typically highly uncertain, often by tens of percentage points.
The respective articles do not address this uncertainty, regularly presenting insignificant figures.
Therefore, their audience is unintentionally mislead  into believing that classifier metrics are known precisely although this is clearly not the case.

We could show that some classifiers carry a non-negligible risk of being deceptive.
Moreover, empirical uncertainties, determined by repeating a classification experiment, would be even larger than the true uncertainty of a metric due to small \gls{n}.
Thus, many published classification metric point estimates are unlikely to be reproducible.

Poorly understood classifiers potentially harm individuals and society.
Our example on cocaine purity analysis has shown that the number of miscarriages of justice due to an insufficiently tested classifier could be alarmingly high.
Similarly, the likelihood of misdiagnoses and subsequent wrongfully administered therapies based on a medical classifier remain obscure unless we account for \acrlong{n}.
In basic science, uncertain classifiers can misguide further research and thus waste resources.
During the identification of molecules with therapeutic potential, a poor classifier would discard the most promising ones or lead the researchers to a dead-end.
Since time and funding are finite, this would decrease progress resulting in economic as well as medical damages.

The example of the Kaggle challenge shed light on the problem of uncertain performance in classifier meta-analysis. There, sample size is usually large but performance differences are minute.
Consequently, classifier or algorithm rankings are uncertain.

We can interpret the frequent failure to account for metric uncertainty in classification as another facet of the current replication crisis, one root cause of which is neglect of uncertainty.~\cite{Gelman2017, Wasserstein2019}
Classifier evaluation should be considered like any other experiment.
It is obvious that a physical quantity cannot be measured exactly, and neither can a classifier metric.
Thus, its uncertainty should be estimated and properly communicated.

For easy access to the method proposed here, we provide a free open-source software at \url{https://github.com/niklastoe/classifier_metric_uncertainty}. The software can be used without programming in an interactive web interface.
The only required input is the \acrlong{cm}, i.e.\ information that is usually available for published classifiers.
The software then computes uncertainty for any of the commonly used classifier metrics.
Moreover, sample sizes that are required to achieve a given exactness of a metric can be estimated according to Equation~\ref{eq:ssd_approximation}.
We hope this contributes to more realistic expectations, more thoughtful allocation of resources and ultimately reliable performance assessments of classifiers.

Our approach can be extended to similar problems.
Multiclass classification can be modeled by $c+1$ multinomial distributions (where $c$ is the number of classes), analogously to Figure~\ref{fig:scheme}.
Another extension of our approach is the computation of error bars of the popular receiver operating characteristic (ROC) curve, which is basically a vector of \glspl{cm}. 
It would be more difficult to use our approach to compute the uncertainty of the area under the ROC curve (AUC), another popular classifier metric. However, the AUC, too, will be uncertain for finite \gls{n}. 
A further extension is the inclusion of classification scores in a distributional model~\cite{Brodersen2010a}, because the scores contain additional information that leads to a better understanding of \gls{mu}.

Our approach only captures the uncertainty arising from finite \gls{n}.
Other sources of uncertainty such as over- or underfitting, data and publication bias etc.\ need to be considered separately.
For instance, comparison of metric posterior distributions calculated separately for the training and test data could help to assess overfitting.
Without such additional analyses, the posterior distributions obtained with our method are probably often too optimistic.

\acknow{We thank Paul B\"urkner, Kai Horny, and Martin Theissen for fruitful discussion.
Funding from Deutsche Forschungsgemeinschaft through project CRC1093/A7 is gratefully acknowledged.}
\showacknow{}

\bibliography{manuscript}{}

\end{document}


\setcounter{table}{0}
\renewcommand{\thetable}{S\arabic{table}}%
\setcounter{figure}{0}
\renewcommand{\thefigure}{S\arabic{figure}}%
\setcounter{equation}{0}
\renewcommand{\theequation}{S\arabic{equation}}%
\setcounter{section}{0}
\renewcommand{\thesection}{S\arabic{section}}%
\setcounter{page}{1}
\renewcommand{\thepage}{S\arabic{page}}%

\maketitle

\section{Priors}
\label{s:metric_priors}

\begin{figure}[!htb]
\centering
\includegraphics[width=6.7in]{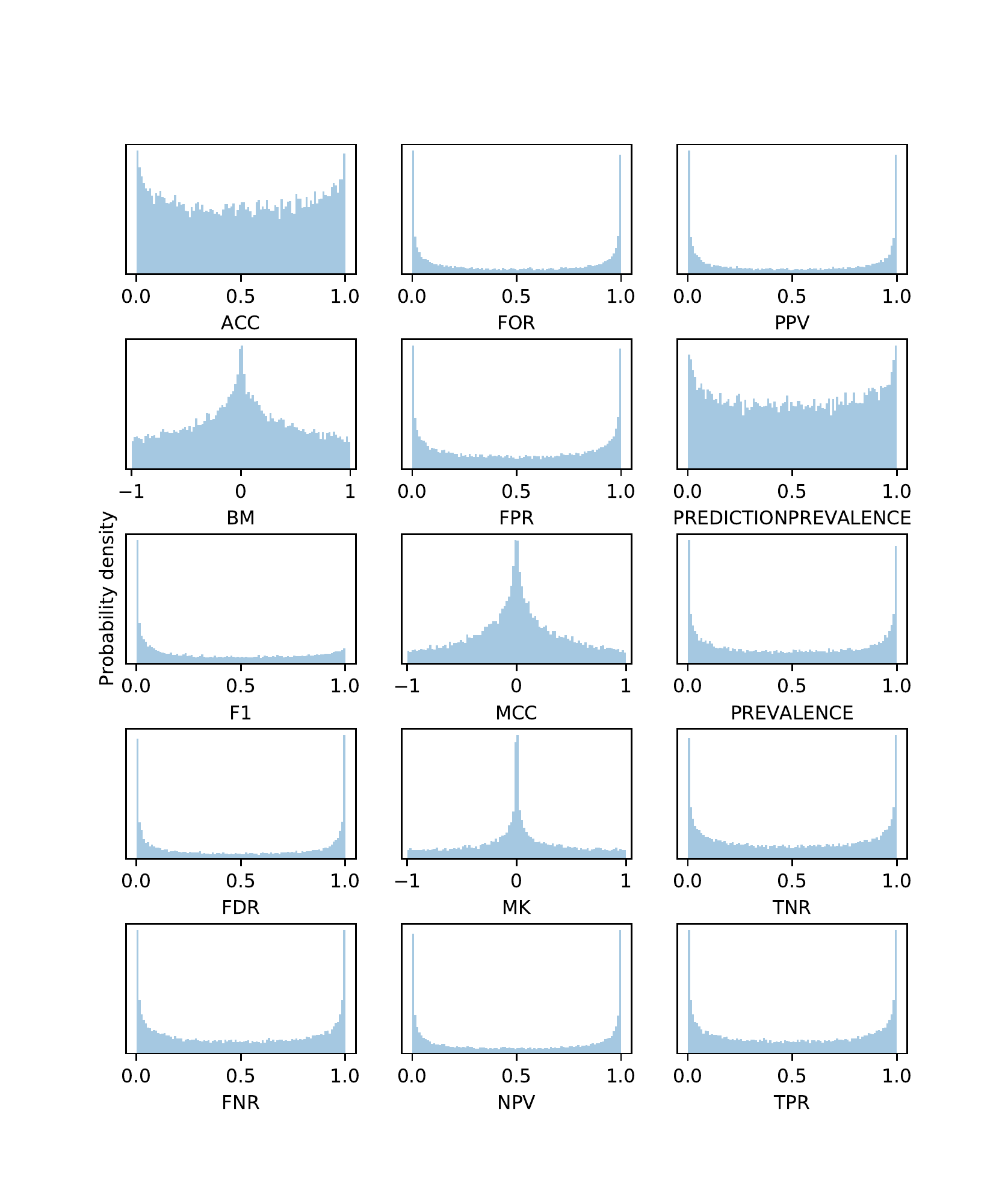}
\caption{Priors on the metrics if Jeffreys priors are used for \gls{prev}, \gls{tpr}, \gls{tnr}}
\label{fig:je_prior_metrics}
\end{figure}

The prior distribution can be interpreted as expression of previous knowledge, which in turn can be expressed in terms of previous observations. In this sense, the Laplace (or flat) prior is equivalent to two previous observations for each \gls{prev}, \gls{tpr} and \gls{tnr}, which is usually a questionable assumption.
Since \gls{n} is small in some of the examples discussed in this study, this assumption could have an impact on the posterior distribution.
Nevertheless, we consider this prior to be the most suitable objective prior.
Haldane's prior, $\mathrm{Beta}(\alpha=0, \beta=0)$, is not adequate since it yields an improper posterior if any entry of the \gls{cm} is zero, which is often the case.
Jeffreys prior, $\mathrm{Beta}(\alpha=0.5, \beta=0.5)$, does not have this problem but leads to implausible U-shaped priors for some metrics (\autoref{fig:je_prior_metrics}).

\begin{figure}[!htb]
\centering
\includegraphics[width=6.7in]{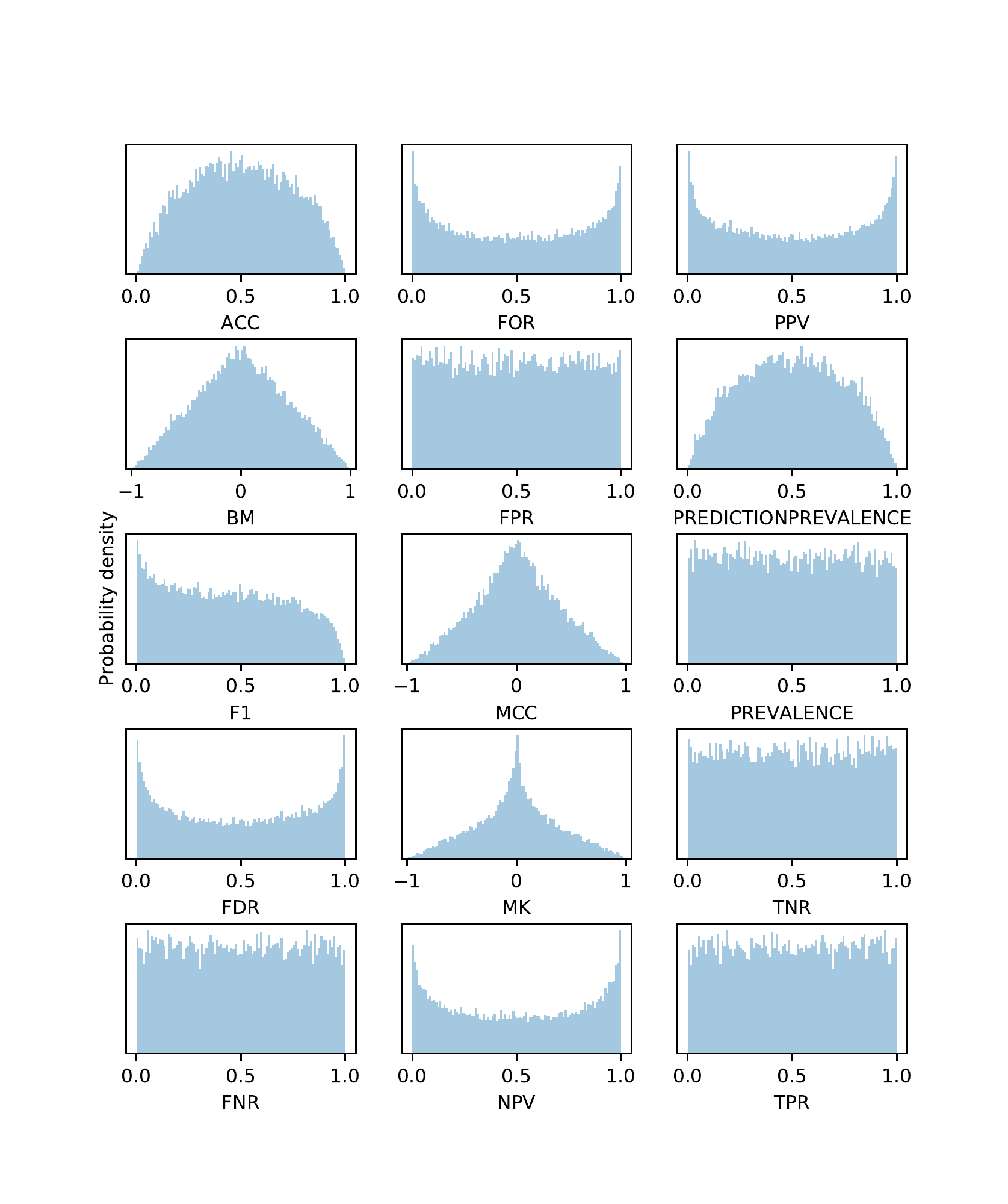}
\caption{Priors on the metrics if Laplace priors are used for \gls{prev}, \gls{tpr}, \gls{tnr}}
\label{fig:bl_prior_metrics}
\end{figure}

\section{Marginals of the \acrlong{cm}}
\label{s:marginals}

There are three scenarios for the marginals of the \gls{cm}.
In principle, the marginals of the columns and rows of the \gls{cm} could both be fixed, which would mean that \gls{prev} and the number of positive/negative predictions are known exactly beforehand.
Fisher's exact test was designed to evaluate whether a binary classifier performs better than random guessing for this specific case.~\cite{Fisher1922}
It remains popular, yet the underlying assumption is usually violated.~\cite{mcelreath2018,Gelman2003}

A fixed \gls{prev} and an unspecified marginal on the predicted labels is more common.
For instance, in a controlled study, test sets may be curated to include 50\% patients suffering from a disease and 50\% healthy subjects in a control group.
In this example there is no uncertainty in \gls{prev}, but it is fixed at \gls{prev}=0.5.

If \gls{prev} in the test set is not deliberately chosen before the compilation, it must be determined from the data set.
For small sample sizes, \gls{prev} is uncertain like all other metrics.
In the present study, we infer \gls{prev} from the \gls{cm} but our method also copes with fixed \gls{prev}.

\clearpage
\section{Literature examples}

\begin{table}[!htb]
\centering
\include{literature_cms}
\caption{Literature examples of classifiers with small \acrfull{n}. Citations were recorded on Google Scholar on June 16th, 2020 at 12:55~pm.}
\label{tab:literature_cms}
\end{table}

\clearpage
\section{Proof that variance of metric distributions calculated from synthetic confusion matrices is systematically too large}
\label{s:iv_converges_to_itheta}

For a \gls{cpm} following a Dirichlet distribution with parameter vector $\alpha$

\begin{equation}
\parm{} \sim \mathrm{Dirichlet}(\alpha)
\end{equation}
where $\alpha$ is the sum of the \gls{cm} and the prior, the expected value and variance are 

\begin{align}
\expected{\parm{i}} &= \frac{\alpha_i}{\alpha_0} \\
\var{\parm{i}} &= \frac{\alpha_i}{\alpha_0} \left( \frac{1 - \frac{\alpha_i}{\alpha_0}}{1 + \alpha_0} \right)
\end{align}
where $\alpha_0 = \sum \alpha_k$. 
The expected value and variance of the entry \cmentry{i} of a confusion matrix generated by a multinomial distribution 

\begin{equation}
V = \left[ \cmentry{\acrshort{tp}}, \cmentry{\acrshort{fn}}, \cmentry{\acrshort{tn}}, \cmentry{\acrshort{fp}} \right] \sim \mathrm{Multinomial}(\parm{}, \acrshort{n})
\end{equation}
is given by 

\begin{align}
\expected{\cmentry{i}} &= \gls{n} \frac{\alpha_i}{\alpha_0} = \gls{n} \expected{\theta_i} \\
\var{\cmentry{i}} &= \gls{n} (\gls{n} + \alpha_0) \frac{\alpha_i}{\alpha_0} \left( \frac{1 - \frac{\alpha_i}{\alpha_0}}{1 + \alpha_0} \right) = \gls{n} (\gls{n} + \alpha_0) \var{\parm{i}}
\end{align}
From this, we can calculate the expected value and variance for the proportion of $i$, $\frac{\cmentry{i}}{\gls{n}}$

\begin{align}
\expected{\frac{\cmentry{i}}{\gls{n}}} &= \frac{1}{\acrshort{n}} \expected{\cmentry{i}} = \expected{\parm{i}} \\
\var{\frac{\cmentry{i}}{\gls{n}}} &= \frac{1}{\gls{n}^2} \var{\cmentry{i}}
= \left( 1 + \frac{\alpha_0}{\gls{n}} \right) \var{\parm{i}}
\label{eq:expected_multinomial}
\end{align}
Whereas \expected{\frac{\cmentry{i}}{\gls{n}}} is independent of \gls{n}, $\var{\frac{\cmentry{i}}{\gls{n}}}$ is not.
In Caelen's approach, $\gls{n} \approx \alpha_0$.
Therefore, the variance will be overestimated by approximately a factor of two. 
Since the variance of $\frac{\cmentry{i}}{\gls{n}}$ are overestimated w.r.t.\ \parm{i}, the same holds for $\frac{\cmentry{}}{\gls{n}}$ w.r.t.\ \parm{} and metrics calculated on $\frac{\cmentry{}}{\gls{n}}$ and \parm{}, respectively.

If \gls{n} was increased beyond $\alpha_0$, it would converge towards the true variance 

\begin{equation}
\lim\limits_{\gls{n} \rightarrow \infty}{\var{\frac{\cmentry{i}}{\gls{n}}}} = \var{\parm{i}}.
\end{equation}

\section{Rule for sample size determination of metrics modeled by a beta distribution}
\label{s:mu_ssd}

For a normal distribution, approximately 95\% of the density are within two standard deviations $\sigma$ from the mean.
Therefore, the length of the 95\% highest posterior density interval will be close to $4 \sigma$.
According to the central limit theorem, beta distributions behave for large sample sizes like normal distributions.
The standard deviation $\sigma$ of a beta distribution is given by

\begin{equation}
\sigma = \sqrt{\frac{\alpha \cdot \beta}{(\alpha + \beta)^2 (\alpha + \beta + 1)}}.
\end{equation}
where $\alpha$ and $\beta$ are the counts of observations per class, where the meaning of ``class'' depends on the studied metric.
As discussed in the main text, if one is looking at \gls{acc}, $\alpha$ denotes correct classifications (\acrshort{tp}~+~\acrshort{tn}) and $\beta$ denotes wrong classifications (\acrshort{fp}~+~\acrshort{fn}).
In the case of \gls{tpr}, $\alpha$ counts the number of \glspl{tp} whereas $\beta$ counts \glspl{fn}.

To make explicit the dependency on sample size \gls{n}, we express $\alpha$ as $a \cdot \acrshort{n}$ and $\beta$ as $b \cdot \acrshort{n}$ with fractions $a=\frac{\alpha}{N}, b=\frac{\beta}{N}$ of the two classes.

\begin{align}
\sigma &= \sqrt{\frac{a \cdot \acrshort{n} \cdot b \cdot \acrshort{n}}{(a \cdot \acrshort{n} + b \cdot \acrshort{n})^2 (a \cdot \acrshort{n} + b \cdot \acrshort{n} + 1)}} \\
\sigma &= \sqrt{\frac{\acrshort{n}^2 \cdot a \cdot b }{\acrshort{n}^2 (a  + b)^2  (\acrshort{n} (a + b)  + 1)}} \\
\sigma &= \sqrt{\frac{a \cdot b }{(a  + b)^2  (\acrshort{n} (a + b)  + 1)}} \label{eq:1}
\end{align}
Since $\alpha + \beta = \acrshort{n}$, we know that $a + b = 1$. 
Now we can simplify Eq (\ref{eq:1}) to

\begin{equation}
\sigma = \sqrt{\frac{a \cdot b }{\acrshort{n}  + 1}}
\end{equation}
For large \acrshort{n}, this approximates to

\begin{equation}
\sigma \approx \sqrt{\frac{a \cdot b }{\acrshort{n}}}
\end{equation}
$\sigma$ is largest if $a=b=0.5$. 

\begin{align}
\sigma_{max} &\approx \sqrt{\frac{0.5 \cdot 0.5}{\acrshort{n}}} \\
\sigma_{max} &\approx \frac{0.5}{\sqrt{\acrshort{n}}}
\end{align}
In the main text, we have defined \gls{mu} as the length of the 95\% highest posterior density interval.
Therefore, its upper limit can be approximated as $4 \sigma \approx \frac{2}{\sqrt{\acrshort{n}}}$.
If one cannot reject the possibility that $a=b=0.5$, one will need $\frac{4}{\acrshort{mu}^2}$ samples to obtain the desired \gls{mu}.

\bibliography{supplementary}{}

%% file: packages.tex
\usepackage{booktabs}
\usepackage{glossaries}
\usepackage{graphicx}
\usepackage{caption}
\usepackage{subcaption}
\usepackage{url}
\usepackage{xr}

%% file: acronyms.tex
\newcommand{\mytitle}{Classifier uncertainty: evidence, potential impact, and probabilistic treatment}

\newcommand{\parm}[1]{\ensuremath{\theta_{#1}}}
\newcommand{\cm}{\ensuremath{V}}

\newcommand{\ensuremathrm}[1]{\ensuremath{\mathrm{#1}}}

\newacronym{cm}{\ensuremathrm{CM}}{confusion matrix}
\newacronym{cpm}{\parm{}}{confusion probability matrix}
\newacronym{n}{\ensuremath{N}}{sample size}

\newacronym[\glslongpluralkey={metric uncertainties}]{mu}{MU}{metric uncertainty}
\newacronym{nhst}{NHST}{null hypothesis significance testing}

\newacronym{tp}{\ensuremathrm{TP}}{true positive}
\newacronym{fp}{\ensuremathrm{FP}}{false positive}
\newacronym{fn}{\ensuremathrm{FN}}{false negative}
\newacronym{tn}{\ensuremathrm{TN}}{true negative}

\newacronym{rinf}{\ensuremath{r_{\mathrm{informative}}}}{chance to be an informative classifier}
\newacronym{rdec}{\ensuremath{r_{\mathrm{deceptive}}}}{chance to be a deceptive classifier}
\glsunset{rinf}
\glsunset{rdec}

\newacronym{mcc}{\ensuremathrm{MCC}}{Matthews correlation coefficient}
\newacronym{bm}{\ensuremathrm{BM}}{bookmaker informedness}
\newacronym{fdr}{\ensuremathrm{FDR}}{false discovery rate}
\newacronym{for}{\ensuremathrm{FOR}}{false omission rate}
\newacronym{prev}{\ensuremath{\phi}}{prevalence}
\newacronym{acc}{\ensuremathrm{ACC}}{accuracy}
\newacronym{tpr}{\ensuremathrm{TPR}}{true positive rate}
\newacronym{tnr}{\ensuremathrm{TNR}}{true negative rate}

\newacronym{bbd}{BBD}{beta-binomial distribution}

%% file: rodrigues_cm.tex
\begin{tabular}{lrr}
\toprule
{} &  r=high &  r=low \\
\midrule
m=high &    26 &    2 \\
m=low &     0 &    6 \\
\bottomrule
\end{tabular}

%% file: literature_cms.tex
\resizebox{\textwidth}{!}{
\begin{tabular}{lllrrrrrr}
\toprule
{} &                              DOI &    Location &   TP &  FN &   TN &  FP &    N &  Citations \\
\midrule
1   &        10.1080/10629360903278800 &     Table 2 &    5 &   0 &    3 &   0 &    8 &         10 \\
2   &                10.1021/ci200579f &     Table 3 &   10 &   0 &    3 &   1 &   14 &         48 \\
3   &                 10.1021/ci020045 &     Table 5 &    6 &   0 &    7 &   1 &   14 &         51 \\
4a  &              10.1155/2015/485864 &     Table 4 &    5 &   1 &   10 &   1 &   17 &         10 \\
4b  &              10.1155/2015/485864 &     Table 5 &    4 &   2 &   10 &   1 &   17 &         10 \\
5a  &     10.1016/j.ejmech.2010.11.029 &     Table 6 &   16 &   1 &    3 &   2 &   22 &         86 \\
5b  &     10.1016/j.ejmech.2010.11.029 &    Table 10 &    8 &   9 &    4 &   1 &   22 &         86 \\
6a  &      10.1016/j.vascn.2014.07.002 &     Table 2 &    2 &  12 &   19 &   1 &   34 &         77 \\
6b  &      10.1016/j.vascn.2014.07.002 &     Table 3 &   10 &   4 &   20 &   0 &   34 &         77 \\
7a  &       10.5935/0103-5053.20130066 &     Table 2 &   26 &   0 &    6 &   2 &   34 &         61 \\
7b  &       10.5935/0103-5053.20130066 &     Table 3 &   24 &   2 &    6 &   2 &   34 &         61 \\
8   &  10.1016/j.scitotenv.2018.05.081 &     Table 2 &   28 &   9 &    3 &   4 &   44 &         18 \\
9a  &          10.4314/wsa.v36i4.58411 &     Table 2 &   19 &   3 &   18 &  10 &   50 &         14 \\
9b  &          10.4314/wsa.v36i4.58411 &     Table 2 &   21 &   1 &   20 &   8 &   50 &         14 \\
10  &       10.1016/j.bspc.2017.01.012 &    Figure 2 &   31 &   5 &   24 &   4 &   64 &         80 \\
11  &               10.1039/C7MD00633K &    Figure 3 &   40 &   7 &   15 &   8 &   70 &          9 \\
12  &         10.3389/fnins.2018.01008 &    Figure 3 &   31 &   9 &   20 &  13 &   73 &          1 \\
13a &       10.4315/0362-028X-61.2.221 &     Table 3 &   79 &  14 &   19 &   0 &  112 &         52 \\
13b &       10.4315/0362-028X-61.2.221 &     Table 3 &   89 &   4 &   16 &   3 &  112 &         52 \\
14a &       10.1016/j.ancr.2014.06.005 &  Figure 6.3 &  136 &   2 &    2 &  12 &  152 &          7 \\
15a &       10.1016/j.saa.2016.09.028  &     Table 2 &    3 &  12 &  150 &   0 &  165 &         65 \\
15b &       10.1016/j.saa.2016.09.028  &     Table 2 &    6 &   9 &  150 &   0 &  165 &         65 \\
16  &     10.1021/acs.analchem.7b00426 &     Table 3 &  188 &   0 &   13 &   2 &  203 &         28 \\
14b &       10.1016/j.ancr.2014.06.005 &     Table 3 &  253 &  27 &   11 &  59 &  350 &          7 \\
\bottomrule
\end{tabular}
}

%% file: manuscript.bbl
\begin{thebibliography}{10}

\bibitem{baker2016reproducibility}
Baker M (2016) {Is there a reproducibility crisis?}
\newblock {\em Nature} 533(26):353--366.

\bibitem{Dietterich1997}
Dietterich TG (1998) {Approximate Statistical Tests for Comparing Supervised
  Classification Learning Algorithms}.
\newblock {\em Neural Comput.} 10(7):1895--1923.

\bibitem{Benavoli2017}
Benavoli A, Corani G, Dem{\v{s}}ar J, Zaffalon M (2017) {Time for a change: A
  tutorial for comparing multiple classifiers through Bayesian analysis}.
\newblock {\em J. Mach. Learn. Res.} 18:1--36.

\bibitem{Brodersen2010}
Brodersen KH, Ong CS, Stephan KE, Buhmann JM (2010) {The balanced accuracy and
  its posterior distribution}.
\newblock {\em Proc. - Int. Conf. Pattern Recognit.} pp. 3121--3124.

\bibitem{Caelen2017}
Caelen O (2017) {A Bayesian interpretation of the confusion matrix}.
\newblock {\em Ann. Math. Artif. Intell.} 81(3-4):429--450.

\bibitem{Kruschke2015c}
Kruschke JK (2015) {Bayes' Rule} in {\em Doing Bayesian Data Anal.}
\newblock (Elsevier), pp. 99--120.

\bibitem{Gelman1992}
Gelman A, Rubin DB (1992) {Inference from Iterative Simulation Using Multiple
  Sequences}.
\newblock {\em Stat. Sci.} 7(4):457--472.

\bibitem{Brooks1998}
Brooks SP, Gelman A (1998) {General Methods for Monitoring Convergence of
  Iterative Simulations}.
\newblock {\em J. Comput. Graph. Stat.} 7(4):434--455.

\bibitem{Salvatier2016}
Salvatier J, Wiecki TV, Fonnesbeck C (2016) {Probabilistic programming in
  Python using PyMC3}.
\newblock {\em PeerJ Comput. Sci.} 2(4):e55.

\bibitem{chicco2017}
Chicco D (2017) {Ten quick tips for machine learning in computational biology}.
\newblock {\em BioData Min.} 10(1):1--17.

\bibitem{Powers2011}
Powers DMW (2011) {Evaluation: From Precision, Recall and F-Measure To Roc,
  Informedness, Markedness {\&} Correlation}.
\newblock {\em J. Mach. Learn. Technol.} 2(1):37--63.

\bibitem{Calvo2019}
Calvo B, et~al. (2019) {Bayesian performance analysis for black-box
  optimization benchmarking} in {\em Proc. Genet. Evol. Comput. Conf. Companion
  - GECCO '19}.
\newblock (ACM Press, New York, New York, USA), pp. 1789--1797.

\bibitem{Kruschke2015l}
Kruschke JK (2015) {\em {Goals, Power, and Sample Size}}.
\newblock pp. 359--398.

\bibitem{Gelman2017}
Gelman A, Carlin J (2017) {Some Natural Solutions to the p -Value Communication
  Problem—and Why They Won't Work}.
\newblock {\em J. Am. Stat. Assoc.} 112(519):899--901.

\bibitem{Wasserstein2019}
Wasserstein RL, Schirm AL, Lazar NA (2019) {Moving to a World Beyond “p
  {\textless} 0.05”}.
\newblock {\em Am. Stat.} 73(sup1):1--19.

\bibitem{Brodersen2010a}
Brodersen KH, Ong CS, Stephany KE, Buhmann JM (2010) {The binormal assumption
  on precision-recall curves}.
\newblock {\em Proc. - Int. Conf. Pattern Recognit.} (Section IV):4263--4266.

\end{thebibliography}


\begin{thebibliography}{1}

\bibitem{Fisher1922}
Fisher R (1922) {On the interpretation of $\chi$2 from contingency tables, and
  the calculation of P}.
\newblock {\em J. R. Stat. Soc.} 85(1):87--94.

\bibitem{mcelreath2018}
McElreath R (2018) {\em {Statistical rethinking: A Bayesian course with
  examples in R and Stan}}.
\newblock (Chapman and Hall/CRC).

\bibitem{Gelman2003}
Gelman A (2003) {A Bayesian Formulation of Exploratory Data Analysis and
  Goodness-of-fit Testing*}.
\newblock {\em Int. Stat. Rev.} 71(2):369--382.

\end{thebibliography}
